# A Dynamic Neural Network Approach to Generating Robot's Novel Actions: A Simulation Experiment

Jungsik Hwang[1,2] and Jun Tani[2]

[1] Korea Advanced Institute of Science and Technology, Daejeon, Korea
[2] Okinawa Institute of Science and Technology, Okinawa, Japan
{jungsik.hwang, tani1216jp}@gmail.com

**Abstract.** In this study, we investigate how a robot can generate novel and creative actions from its own experience of learning basic actions. Inspired by a machine learning approach to computational creativity, we propose a dynamic neural network model that can learn and generate robot's actions. We conducted a set of simulation experiments with a humanoid robot. The results showed that the proposed model was able to learn the basic actions and also to generate novel actions by modulating and combining those learned actions. The analysis on the neural activities illustrated that the ability to generate creative actions emerged from the model's nonlinear memory structure self-organized during training. The results also showed that the different way of learning the basic actions induced the self-organization of the memory structure with the different characteristics, resulting in the generation of different levels of creative actions. Our approach can be utilized in human-robot interaction in which a user can interactively explore the robot's memory to control its behavior and also discover other novel actions.

## I. Introduction

Imagine an entertaining robot which can learn actions from its user. If the robot is only capable of reproducing the behaviors that it has learned, the user might easily lose his/her interests in the interaction with the robot. In addition, it is cumbersome for the user to teach every single behavior of the robot. Therefore, it would be desirable if the robot can generate not only the learned behaviors but also novel ones by combining and manipulating the already learned actions.

In this study, we employ a dynamic neural network approach to investigate how a humanoid robot can generate novel and creative actions from its own experience of learning basic actions. A machine learning approach to creativity has been one of the emerging topics [1-4] and it provides a suitable platform to investigate creativity "in relation with knowledge" [5, 6]. We assume that robot's novel actions may emerge while combining and manipulating the basic actions that are learned previously and stored in the robot's memory.

To examine our approach, we propose a dynamic neural network model and conduct a set of simulation experiments using a humanoid robot. The dynamic neural network model in this study is composed of the two modules: an action encoding module and an action generation module. The action encoding module encodes the robot's high-dimensional sequential actions in the low-dimensional continuous space. The action generation module learns the robot's sequential actions in a hierarchical structure and generates them in accordance with the action encoding module. During the training process, the model learns basic actions in a holistic manner. That is, the action generation module learns the body dynamics while the action encoding module learns to map robot's high-dimensional actions into the low-dimensional space. After the training process, the model generates the learned actions as well as the novel actions through modulation and combination of the learned actions. This approach can be seen from the dynamic memory perspective. In [5], it was argued that the source of novelty is the nonlinear memory dynamics of the network. In our study, the model self-organizes the nonlinear memory structure in the action encoding module. Consequently, the proposed model can generate not only learned behaviors but also the novel behaviors.

There have been several studies on generating robot's novels actions using machine learning techniques [5, 7, 8]. In [5], the generation of creative goal-directed behaviors by exploiting cortical chaos was investigated. They showed that novel actions can be generated from the memory dynamics self-organized from the consolidative learning of the exemplars. In [7], the cognitive architecture for artificial creativity was proposed and it was used to generate humanoid robot's dance movements. Recently, Augello, et al. [8] showed how a robot can generate novel dance behaviors using the variational encoder. In line with those previous studies, this study aims to investigate the generation of novel action from a dynamic neural network perspective.

This work was supported by the Industrial Strategic Technology Development Program (10044009) funded by the Ministry of Knowledge Economy in Korea and Okinawa Institute of Science and Technology Graduate University in Japan.

Jungsik Hwang is with the school of Electrical Engineering in Korea Advanced Institute of Science and Technology (KAIST), Daejeon, South Korea and Okinawa Institute of Science and Technology (OIST), Okinawa, Japan (e-mail: jungsik.hwang@gmail.com). Jun Tani is a corresponding author and he is with OIST (e-mail: tani1216jp@gmail.com).

## II. Dynamic Neural Network Model

*A. Overview*

The proposed model consists of the two modules: the action encoding module and the action generation module (Fig. 1). The action encoding module encodes the robot's actions in the continuous low-dimensional spaces. The action generation module generates the action encoded in the action encoding module. The similar architecture was introduced in [9] where the two different modules encoded the body dynamics and the tool dynamics features respectively.

There are several key features of the proposed model. First, the model encodes robot's high-dimensional actions in the low-dimensional continuous spaces in the action encoding module. Instead of categorizing robot's action discretely, the proposed model encodes the actions in the continuous space without any human intervention, resulting in more flexible generation of robot's action. Second, the model is able to proactively generate robot's action with given action encoding values through the top-down process, without any external inputs. Consequently, the model can mentally anticipate the possible incoming signals without any external information from the environment (i.e. mental simulation) [10, 11].

*B. Action Encoding Module*

The action encoding module maps the robot's high-dimensional actions in the low-dimensional continuous space. The action encoding module is composed of a set of parametric bias (PB) neurons introduced in [12-14]. Usually, the PB is a static vector input to the model and it acts as the bifurcation parameter [13]. Previous studies [12-14] showed that robot's continuous actions can be encoded in the PB space. In our model, the PB is static vector input to the action generation module. That is, the action encoding module is unidirectionally connected to each layer in the action generation module. The neurons in each layer (except $P_I$ and $P_O$ layer) of the action generation module receive inputs from the action encoding module. The activation value of the $j$th neuron in the action encoding module is computed as follow:

$$P_j = \tanh(\rho_j) \tag{1}$$

where $\rho_j$ is the internal state of the $j$th PB neuron. $\rho_j$ is obtained from the learning process in the process of minimizing the training error. In other words, the PB vector $\rho_j$ is handled as a learnable parameter of the model during training (See Section II.D also). Then, the model can generate different actions by feeding different PB values in the action encoding module. Note that $\rho_j$ is obtained for each training pattern.

*C. Action Generation Module*

The action generation module generates the robot's sequential actions based on the given PB values ($P$) from the action encoding module. The input and the output of the model is a proprioceptive signal (i.e. joint position values). At each time step, the proprioceptive signal is given to the input of the model and then, the model predicts the proprioceptive signal in the next time step.

A hierarchical continuous time recurrent neural network called multiple timescales recurrent neural network (MTRNN) [15] was used to implement the action generation module. In general, an MTRNN model consists of a set of layers exhibiting different timescales dynamics. This characteristic can be achieved by imposing different temporal constraints on each layer of the MTRNN model. Previous studies have [15-17] shown that MTRNN can decompose the robot's action into the primitives and compose them in a hierarchical structure.

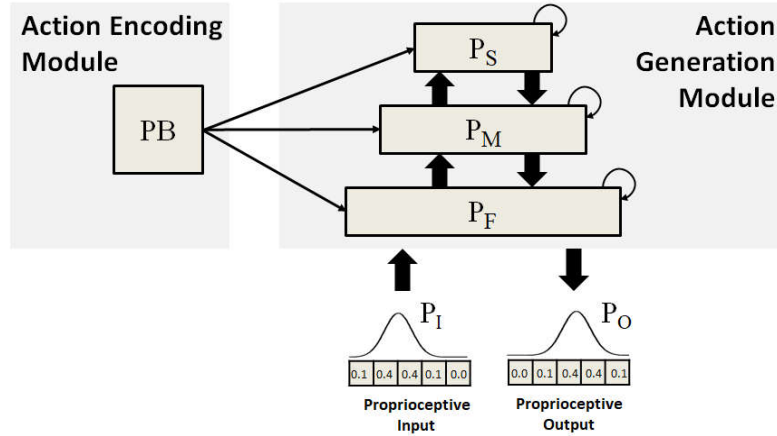

Fig. 1. The proposed neural network model consists of the action encoding module (left) and the action generation module (right). The action generation module consists of Proprioception Input ($P_I$), Proprioception Output ($P_O$), Proprioception Fast ($P_F$), Proprioception Middle ($P_M$) and Proprioception Slow ($P_S$).

In our study, five MTRNN layers are implemented: Proprioception Input and Output ($P_I$, $P_O$), Proprioception Fast ($P_F$), Proprioception Middle ($P_M$) and Proprioception Slow ($P_S$) layers. Those layers are bidirectionally connected to the neighboring layers. Also, the neurons in each layer have the recurrent connections to the neurons in the same layer. The different temporal constraints are imposed on each MTRNN layer to achieve different timescales dynamics. Particularly, the progressively larger time constants are assigned from the lower-level layers to the higher-level layers as suggested in the previous studies [15-17]. Consequently, the neurons in the higher-level layer with the larger time constants exhibit relatively slower dynamics compared to the ones in the lower-level layers.

The internal states $u$ and the activation $y$ of the $i$th neuron at each time step $t$ in the action generation module are computed as follow:

$$u_i^t = \left(1 - \frac{1}{\tau_i}\right) u_i^{t-1} + \frac{1}{\tau_i} \left( \sum_{j \in PB} w_{ij} P_j + \sum_{k \in MTRNN} w_{ik} y_k^{t-1} + b_i \right) \quad (2)$$

$$y_i^t = \begin{cases} \frac{\exp(u_i^t)}{\sum_{j \in P_O} \exp(u_j^t)} & if \ i \in P_O \\ \tanh(u_i^t) & otherwise \end{cases} \quad (3)$$

$\tau$ is the time constant, $j$ denotes the PB neuron in the action encoding module and $k$ denotes the neurons in the neighboring MTRNN layers as well as the current layer in the action generation module. $w_{ij}$ is the connection weight from the $j$th neuron in the action encoding module to the $i$th neuron in action generation module, b is the bias, $w_{ik}$ is the connection weight from the $k$th neuron to the $i$th neuron in action generation module, and P is the activation value of PB in the action encoding module. The softmax activation function was used in the output layer ($P_O$) and hyperbolic tangent was used elsewhere.

*D. Training the Model*

The model is trained in a supervised manner in which the dynamic proprioceptive patterns are provided as a teaching signal. The training data can be obtained from the kinesthetic teaching [18, 19] process where the robot is manually operated by the experimenter. We train the model to generate one-step look-ahead prediction using backpropagation through time (BPTT) [20]. During the training process, the training error $E$ is defined as the discrepancy between the generated signal ($y$) and the teaching signal ($\bar{y}$) and it is represented by the Kullback-Leibler divergence between those signals (4).

$$E = \sum_t \sum_i \bar{y}_i^t \log \frac{\bar{y}_i^t}{y_i^t} \quad (4)$$

Where $i$ and $t$ denote the softmax neuron in $P_O$ and the time step respectively. Then, the model's learnable parameters, such as weights ($w$), biases ($b$) and the internal states of the action encoding module ($\rho$) are optimized to minimize the error with the learning rate $\eta$. Note that the internal states of the neurons in the action encoding module ($\rho$) are considered as learnable variables, meaning that those values are obtained during the training in the process of minimizing the training error $E$. The model's learnable parameters including weights and biases are same for all the training patterns whereas the PB values are obtained separately for each training pattern.

In our study, three different types of learning method are compared [10]. The first method is called open-loop training. In this method, the input to the model $y_i^t$ is obtained from the training dataset ($y^*$) and this can be done by setting the closed-loop ratio $\gamma$ as 0.0 in (5). The second method is called closed-loop training. In this method, $\gamma$ is set to 1.0, meaning that the prediction output of the model ($\tilde{y}$) is fed back into the input of the model in the next time step. Consequently, the model is able to generate robot's action without external inputs. Assuming that the model's output as the proprioceptive signals, this process can be considered as the mental simulation of action where the robot anticipates its own actions [11, 16, 17]. The third method is called half closed-loop training. In this method, $\gamma$ is set to 0.5. Consequently, both training data and the prediction output of the model in the previous time step are used as an input to the model at the current time step.

$$y_i^t = \gamma \tilde{y}_i^t + (1 - \gamma) y^{*t}_i \qquad i \in P_I \quad (5)$$

## III. METHOD

*A. Robotic Platform*

A simulated humanoid robot NAO was used in this study (Fig. 2). NAO has 25 degrees of freedom distributed on its head, arms, hands and legs. In our study, eight joints in the arms (right shoulder pitch, right shoulder roll, right elbow yaw, right elbow roll, left shoulder pitch, left shoulder roll, left elbow yaw and left elbow roll) were used to generate robot's action.

The robot was first tutored to generate six basic actions that can be found in boxing such as left jab, right straight, left hook right hook, left uppercut and right uppercut (See the supplementary video). During the tutoring process, the real NAO robot was used to obtain the training data. The experimenter grasped the robot's arms and manually operated the robot to demonstrate such actions (i.e. kinesthetic teaching [18, 19]). Then, the position values of the robot's joints were collected at each time step.

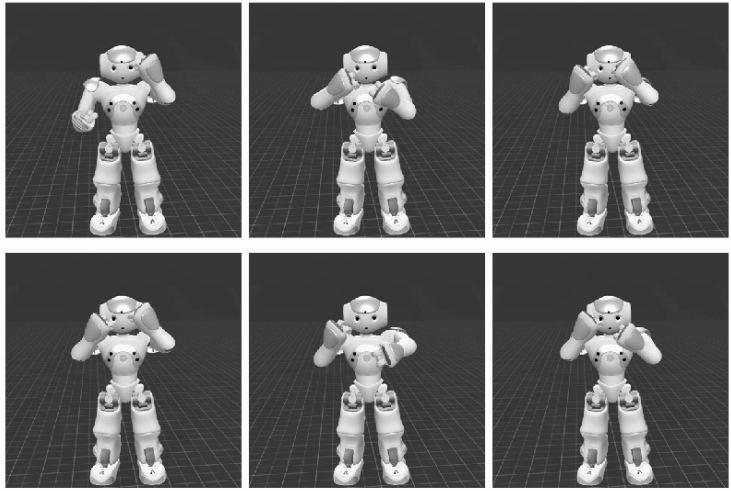

Fig. 2. The simulated NAO robot used in the experiment. The robot was trained to generate six basic actions that can be found in boxing such as left jab, right straight, left hook right hook, left uppercut and right uppercut (See the supplementary video).

*B. Network Settings*

There were 40, 20 and 10 neurons in $P_F$, $P_M$ and $P_S$ layers respectively. The size of the action encoding module (the number of the PB nodes) was limited to two dimensions. Although the size of the dimension is low, an infinite number of robot's actions can be embedded in the continuous PB space. In addition, two-dimensional spaces can be visualized easily, so that the user can easily understand the robot's actions mapped in the action encoding module. There were 80 neurons in $P_I$ and $P_O$ layers, representing 8 groups of joint position value each represented by 10 softmax neurons (See Section III.C for details).

Regarding the time constant settings, progressively larger time constants were assigned from the lower-level layers to the higher-levels layers as suggested in the previous studies [15]. To be more specific, we assigned the 2, 4 and 8 on $P_F$, $P_M$ and $P_S$ layer respectively. There is no time constant for the PB neurons in the action encoding module and the values of the PB nodes did not change (static) while generating the actions (1). The time constant of the $P_I$ and $P_O$ layers were set to 1. These values of the hyper-parameters were found empirically in the preliminary experiments.

*C. Softmax Transformation*

To enhance learning, the softmax transform [10] was used in our study. That is, instead of using the joint position values in radian (analog form), we converted them to the sparse forms. For example, one-dimensional joint position value in radian can be converted into the sparse form using (6).

$$softmax_j = \frac{\exp\left(\frac{-\|reference_j - analog\|^2}{\sigma}\right)}{\sum_{j \in J} \exp\left(\frac{-\|reference_j - analog\|^2}{\sigma}\right)} \quad (6)$$

$softmax_j$ refers to the non-negative value of the *j*th softmax neuron. J is the number of softmax units, indicating how many softmax neurons are used to represent one-dimensional analog value (*analog*). $ref_j$ is the value of the *j*th reference point. The reference points are linearly spaced vector between the minimum and maximum of *analog* in the training dataset. $\sigma$ is another parameter determining a shape of the transformed value, and it is chosen empirically to minimize the conversion error during the softmax transformation.

In our experiment, J was set to 10, meaning that each joint position value was represented by 10 non-negative softmax values. Consequently, there were eight groups of sparse representations of the joint position values where each group is composed of 10 softmax neurons. In our study, $\sigma$ was set to 0.5. During testing, the model's output in the softmax form was converted to radian using (7)

$$analog = \sum_{j \in J} softmax_j \times reference_j \quad (7)$$

*D. Training and Action Generation*

During training, the model was trained to generate the six different basic actions. The model was trained on Tensorflow [21] using the ADAM optimizer [22] and it was trained for 100,000 epochs with the initial learning rate of 0.001. The learnable parameters, such as weights (*w*), biases (*b*) and parametric biases (*ρ*) were initialized with the neutral values at the beginning of training. Note that the different PB values were obtained for each training pattern whereas the other parameters were same across the training patterns. In our experiment, we compared the three conditions with the different closed-loop ratio (γ) during the training process: 0.0 (open-loop), 0.5 (half closed-loop) and 1.0 (closed-loop).

During testing, the training epoch with the lowest training error was selected and the robot's actions were generated by setting the PB values. Specifically, each PB neuron consisted of 200 linearly spaced values between -1 and 1. As the action encoding module was composed of 2 PB neurons, a total number of 40,000 PB values were prepared. Then, the model generated the robot's actions (N = 40,000) with the given PB values in the closed-loop method (i.e. mental simulation).

*E. Measures*

There is no well-grounded measure of creative action generation of robots. Instead of qualitatively analyzing the level of creativeness and novelty of individual action generated by the model, we focused on three aspects of the generated action: appropriateness, novelty, and diversity.

First, we examined whether the generated patterns were appropriate to be used for the robot's action (**Appropriateness**). For instance, if the joint position values change too quickly (Fig. 3 (b). left), they cannot be used because it may harm the actuators in the robot. Also if the joint position values do not change over time (Fig. 3 (b). right), they cannot be used to generate 'action'. Therefore, those 'inappropriate' patterns were filtered out and we calculated the ratio of the 'appropriate' patterns among the entire generated patterns (N = 40,000). To filter out highly fluctuating patterns, we limited the maximum joint angular velocity to be 150% of that of the basic actions in the training data.

Second, we measured the signal similarity between the generated patterns and the training patterns (**novelty**). Intuitively, if a generated pattern differs significantly from the learned patterns, it might be considered as novel. To be more specific, we measured the minimum dynamic time warping (DTW) distance between the generated pattern and the learned patterns. Then, the average of those minimum DTW distances was computed. Therefore, if the average of the minimum DTW distance between the generated pattern and the training patterns is huge, it implies that the generated patterns are generally different from the learned patterns. The novelty was computed iteratively (30 times) with randomly selecting 30 generated patterns at each iteration.

Third, the signal similarity between the generated patterns (**diversity**) was measured. Even if the generated patterns are different from the learned patterns, it would be more desirable if the generated patterns are different from each other. Therefore, we measured the average DTW distance between the generated patterns. Therefore, if the average DTW distance among the generated patterns is huge, it implies that the diverse actions are generated by the model. The diversity was also computed iteratively (30 times) with randomly selecting 30 generated patterns at each iteration.

## IV. RESULTS

The learning converged in all conditions (Fig. 2), meaning that the model was able to learn the six basic patterns. In general, the open-loop training method (γ=0.0) converged faster than the other methods. Under the closed-loop training method (γ=1.0), a bit more fluctuating learning curve was observed.

Table I illustrates the experiment results in terms of three measures used in this study. Regarding appropriateness, the ratio of appropriate patterns among the entire generated pattern (N = 40,000) is reported (*subtotal*). In order to understand what sorts of patterns were generated, we additionally classified those appropriate patterns into the two categories: unlearned and learned actions. If the minimum DTW distance between the generated pattern and the training patterns was higher than the threshold (10.0), the generated pattern was considered the unlearned action. Otherwise, the generated pattern was considered learned action (i.e. training data).

It was observed that the most of the generated patterns were considered appropriate (97.97%) under the closed-loop condition (γ=1.0). In other words, the majority of the patterns generated by the model can be used. In other conditions, the percentage of the appropriate pattern decreased to 82%. However, the classifying the appropriate patterns into two categories showed an interesting result. In the closed-loop condition, about 40% of the appropriate patterns were similar to the training patterns. On the other hand, only 7.58% of the appropriate patterns in the half closed-loop condition (γ=0.5) were considered as the training pattern. This implies that the model trained under the closed-loop manner can reproduce what it has learned well, but it might be less 'creative' than the model trained under the half closed-loop method.

The similar finding was also observed in the measure of novelty. The half closed-loop condition showed the highest degree of novelty (31.71). This suggests that the generated patterns in this condition are more different from the training actions than ones in the other conditions. On the other hand, the lowest level of novelty was observed (18.53) in the closed-loop condition, implying that the generated patterns are more similar to the learned actions than ones in the other conditions. This finding is in line with the measure of appropriateness.

The similar trend was also observed in the measure of diversity. The half-closed loop condition showed the highest level of diversity (48.03) whereas the closed-loop condition showed the lowest level of diversity (35.96). That is, the patterns generated in the half-closed loop condition were different from each other.

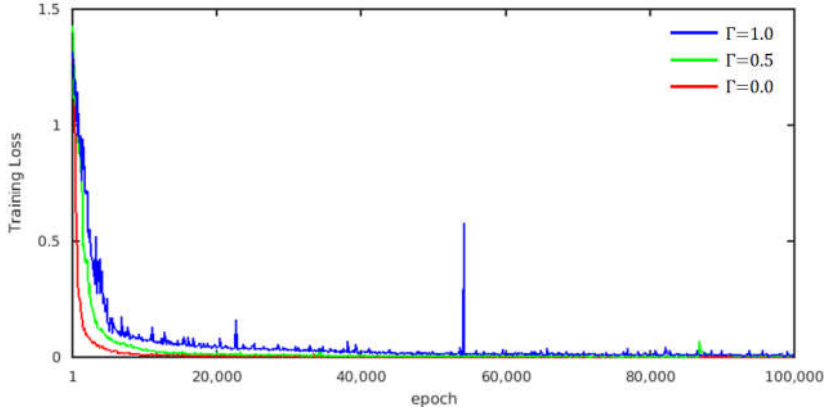

Fig. 2. The learning curve. The model was trained for 100,000 epochs under the three different conditions: (from the top) closed-loop (γ=1.0, blue), half closed-loop (γ=0.5, green) and open-loop (γ=0.0, red).

TABLE I. THE EXPERIMENT RESULT

|  |  | Closed-loop Ratio (γ) during Training | | |
|---|---|---|---|---|
|  |  | 0.0 | 0.5 | 1.0 |
| Appropriateness (%) | Unlearned | 72.21 | 75.26 | 57.95 |
|  | Learned | 11.23 | 7.58 | 40.02 |
|  | Subtotal | 83.44 | 82.84 | 97.97 |
| Novelty |  | 26.02 | 31.71 | 18.53 |
| Diversity |  | 43.12 | 48.03 | 35.96 |

In sum, the result revealed that the most creative patterns were generated in the half-closed loop training condition. The closed-loop condition elicited stable performance where almost all generated patterns were considered appropriate, but those patterns were similar to the training patterns. In contrast, the model trained in the half-closed loop condition generated not only appropriate patterns but also novel and diverse patterns.

This result indicates that the learning method had an important impact on the model's capability of generating creative robot actions. As similar to [5], we assumed that the source of novelty in our model was the nonlinear memory dynamics, particularly in the action encoding module (PB). We hypothesized that the different memory structure had been self-organized during training depending on the condition. Therefore, we additionally analyzed the model's internal memory structure in the action encoding module.

*A. Internal Structure in the Action Encoding Module*

In order to investigate the internal memory structure self-organized in the action encoding module, we visualized the PB spaces (Fig. 4). The colors denote the corresponding basic actions used in training and the contrasts indicate the levels of similarity (high contrast for a higher level of similarity). The X and the Y axes indicate the first and the second neurons in the action encoding module (PB nodes).

In all three conditions, the clusters each encoding specific type of the basic action were observed. That is, each basic action was encoded in a distinct region of the PB space in the action encoding module, and the nearby PB space produced similar patterns (See the supplementary video).

It was observed that the memory structure became simpler when the closed-loop ratio (γ) increased. In the closed-loop condition (γ=1.0), relatively huge spaces encoded the six basic actions. In contrast, in the open-loop condition (γ=0.0), the size of the regions encoding the basic actions decreased. Moreover, the size of those regions was uneven among the basic patterns. For instance, the size of the regions that generated left and right uppercut (L.Upper and R.Upper) were much smaller than the size of the region encoding left hook (L.Hook). With this memory structure, it might be difficult for the model to generate such actions. In the half closed-loop condition (γ=0.5), the size of the region encoding right straight (R.Straight) was bigger than those of the other regions, but the size of the regions that encoded the basic action was similar in general.

As can be seen from Fig. 4, the relatively huge amount of the PB space in the closed-loop condition was occupied by the six clusters encoding the training pattern (40.02%). Therefore, the model in this condition was able to reproduce the learned pattern well, but it was less likely to generate novel actions (See Table I). In the half closed-loop condition, the regions encoding the basic actions occupy the smaller amount of the PB space (7.58%). Consequently, the model trained in this condition was able to generate more diverse and novel actions.

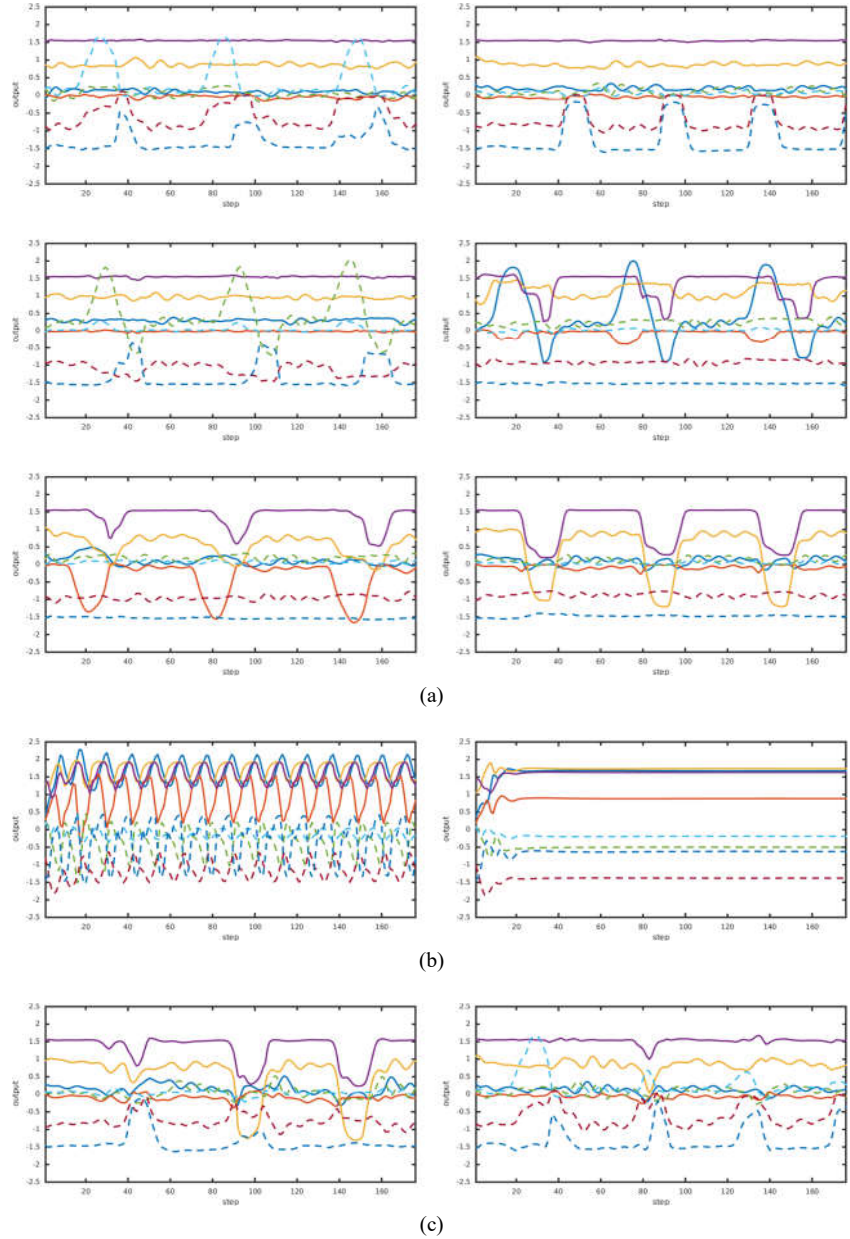

Fig. 3. The patterns generated in the half-closed loop condition. The X axis represents the time steps and the Y axis represents the eight joint position values in radian. The solid and dashed lines represent the joint position values of the right and left arm respectively. (a) Six basic actions used in training. Top row: Left Hook (left) and Left Jab (right). Middle row: Left Uppercut (left) and Right Uppercut (right). Bottom row: Right Hook (left) and Right Straight (right). (b) Example of inappropriate actions filtered out in our result. Highly fluctuating (left) and non-moving (left) patterns. (c) Example of novel actions. A combination of right straight and right uppercut (left) and a combination of left jab and right straight (right). Note that those basic actions in the novel actions are slightly different from the training pattern in (a).

In addition, the analysis revealed a quite 'rugged' landscape of the PB space (Fig. 5). This resulted in abrupt changes in robot's action with a small change in the PB values. As the values of PB changed, the actions generated by the network changed. For instance, Fig. 5 (a) ~ (c) show a transition of an action from right straight (R.Straight) to left jab (L.Jab). As can be seen from the figure, the model's output totally changed to another action with different PB values. Also, it was found that the region between R.Straight and L.Jab encoded the combination of those two actions (Fig. 5 (b)). Sometimes, small changes in PB induced big differences in generated actions (See the supplementary video). These results imply that the model has a nonlinear and complex memory structure. As argued in [5], this highly nonlinear landscape of the PB space is assumed to be the source of 'creative' robot's action in our model. In [5], it was shown that the memory structure could become highly nonlinear when the number of training patterns increased. The findings in this study suggest that the memory structure can be highly nonlinear depending on the learning method as well, even with a smaller number of training data.

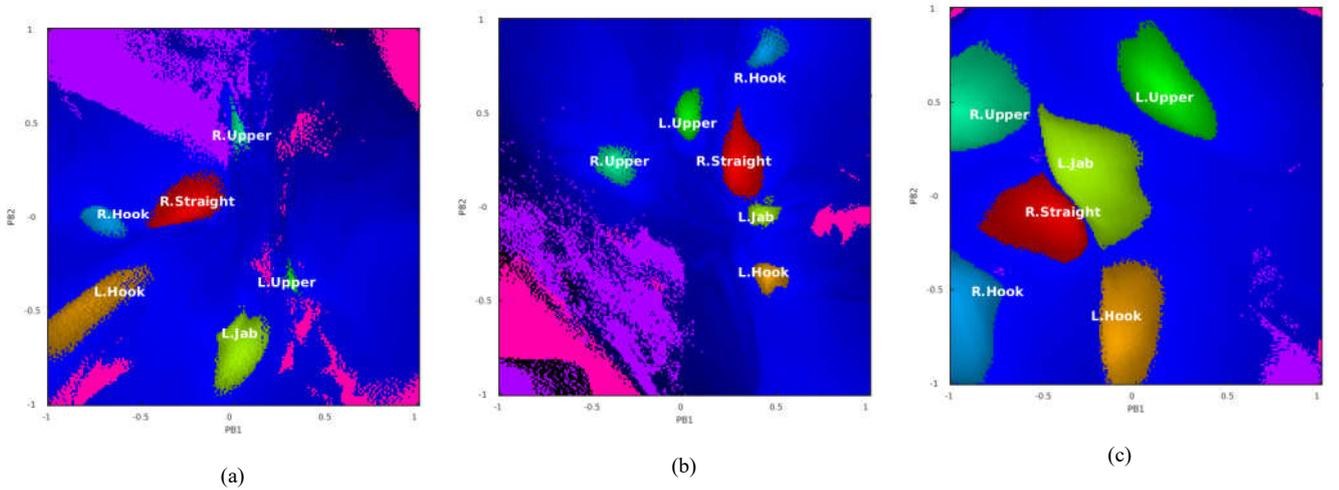

Fig. 4. The PB space in the three different training condition: (a) Open-loop ($\gamma = 0.0$) (b) half closed-loop ($\gamma = 0.5$) and (c) closed-loop ($\gamma = 1.0$). The X and Y axes represent the two PB nodes in the action encoding module. The value of each PB node varies from -1 to 1 with the step size of 0.01. Consequently, the PB space was visualized with 200 × 200 generated patterns (i.e. resolution). The colors and the texts represent the type of basic action used in training. The regions in purple and in pink encode highly fluctuating and non-moving patterns respectively and they were filtered out in our results (See the supplementary video also).

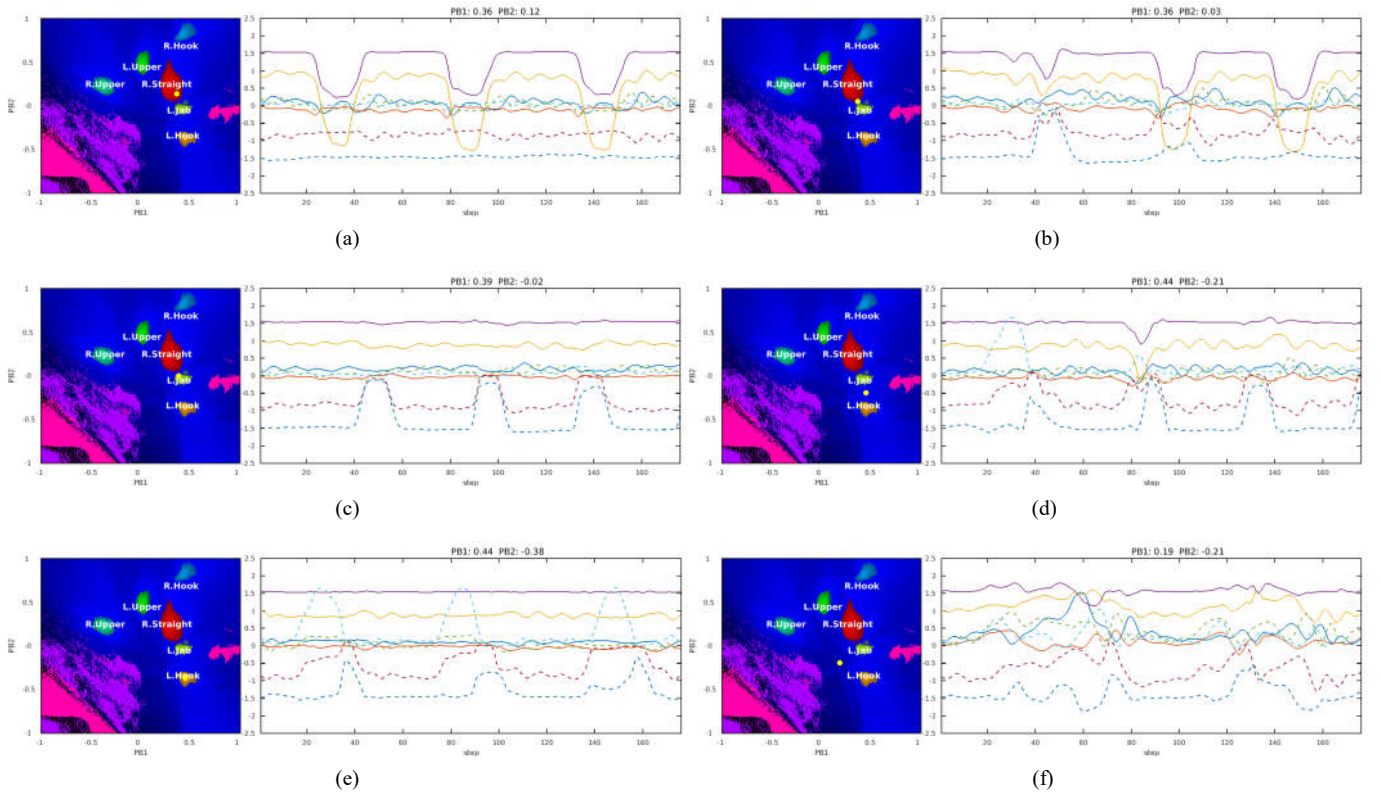

Fig. 5. The model's output with given PB values in the half closed-loop ($\gamma = 0.5$) condition. (a) PB1: 0.36, PB2: 0.12 (b) PB1: 0.36, PB2: 0.03 (c) PB1: 0.39, PB2: -0.02 (d) PB1: 0.44, PB2: -0.21 (e) PB1: 0.44, PB2: -0.38 (f) PB1: 0.19, PB2: -0.21. In each subplot, the PB space (left) and the model's output (right) are depicted. The yellow point in the PB space denotes the location of the current PB value.

## V. Conclusion

In this study, we investigated how a robot can generate novel actions from its own experience of learning basic actions. We proposed a dynamic neural network model which could encode robot's actions in its own memory and reproduce them. The results showed that the proposed model was not only able to reproduce what it had learned but also to generate novel and creative actions through modulating and combining those learned actions. The analysis of the internal memory structure illustrated that the ability to generate novel actions emerged from the nonlinear memory structure self-organized during training in the action encoding module. It was found that the different way of learning the basic actions induced the self-organization of the memory structure with the different characteristics, resulting in the generation of different level of creative actions.

Although this study was conducted in a simulation environment, the method of visualizing the robot's memory structure and generating actions can be implemented in the real robot setting. Consequently, the proposed approach to novel action generation can be utilized in human-robot interaction in which a user can interactively explore the robot's memory to control the robot's behavior and also to discover novel actions.

## Appendix

Supplementary videos are available at https://youtu.be/xsJnJb5zfJ0